\documentclass[manuscript, screen, nonacm]{acmart}

\begin{document}

\title{A conceptual model for leaving the data-centric approach in machine learning}


\author{Sebastian Scher}
\email{sscher@know-center.at}
\affiliation{
 \institution{Know-Center}
 \streetaddress{Sandgasse 36}
 \city{Graz}
 \country{Austria}}
\author{Bernhard Geiger}
\affiliation{
 \institution{Know-Center}
 \streetaddress{Sandgasse 36}
 \city{Graz}
 \country{Austria}}
\author{Simone Kopeinik}
\affiliation{
 \institution{Know-Center}
 \streetaddress{Sandgasse 36}
 \city{Graz}
 \country{Austria}}
\author{Andreas Tr\"ugler}
\affiliation{
 \institution{Know-Center}
 \streetaddress{Sandgasse 36}
 \city{Graz}
 \country{Austria}}
  \affiliation{
 \institution{Institute of Interactive Systems and Data Science, Graz University of Technology}
 \streetaddress{Inffeldgasse 16C}
 \city{Graz}
 \country{Austria}}
\affiliation{
  \institution{Department of Geography and Regional Science, University of Graz}
 \streetaddress{Heinrichstraße 36}
 \city{Graz}
 \country{Austria}}
\author{Dominik Kowald}
\affiliation{
 \institution{Know-Center}
 \streetaddress{Sandgasse 36}
 \city{Graz}
 \country{Austria}}
  \affiliation{
 \institution{Institute of Interactive Systems and Data Science, Graz University of Technology}
 \streetaddress{Inffeldgasse 16C}
 \city{Graz}
 \country{Austria}}

\begin{abstract}
  For a long time, machine learning (ML) has been seen as the abstract problem of learning relationships from data independent of the surrounding settings. This has recently been challenged, and methods have been proposed to include external constraints in the machine learning models. These methods usually come from application-specific fields, such as de-biasing algorithms in the field of fairness in ML or physical constraints in the fields of physics and engineering. In this paper, we present and discuss a conceptual high-level model that unifies these approaches in a common language. We hope that this will enable and foster exchange between the different fields and their different methods for including external constraints into ML models, and thus leaving purely data-centric approaches.
\end{abstract}

\maketitle
\section{Introduction}
The essence of supervised Machine Learning (ML) is to learn the relationship in paired data. For a long time, the development of ML models has been seen as an abstract problem that can rely on data alone, and is agnostic of the surrounding setting. However, in many settings, the predictions made by models that solely rely on data are unsatisfactory in practical applications. Here, we on purpose use the generic word ``unsatisfactory'' instead of more technical terms such as ``not accurate'', as such terms are usually associated with being computed solely from the data. Cases of predictions being unsatisfactory could for example be unfair predictions ---e.g. predictions that are biased towards a specific gender---, or predictions that violate physical laws.
In order to make the predictions satisfactory, one can, in addition to the data, introduce external constraint(s). Here we will deal with this idea on a very high conceptual level that is application- and field-agnostic, and thus can be  applied to a wide range of settings. This in contrast to more technical discussions that focus on a particular problem set (e.g., including knowledge about physics in a physics setting).

We develop a conceptual model for the purpose of ensuring that an ML model adheres to constraints that are not (or not necessarily) represented by the data from which the ML model is learned. 
A related approach is ``Seldonian ML''  \citep{thomas_preventing_2019},  which is a way to include probabilistic constraints into ML applications in order to prevent ``undesirable behaviour''. The approach of Seldonian ML is generic, and can be used for different domains. It can be seen as a mathematical framework that combines requirements for the algorithm specified by the algorithm designer with constraints specified by the user of the algorithm. Our paper employs a view on a much higher, non-mathematical, more conceptual level.

In the first part of the paper, we develop and present the conceptual model. In the second part, we show how it can be applied in two different fields, for both of which there is a large body of research on ensuring that the ML models adheres to certain constraints: physics-informed ML on the one hand, and fair ML on the other hand. While the constraints in these two fields come from very different perspectives ---physical laws in the case of physics-informed ML, and ethical constraints in the case of fair ML --- we show that with our conceptual model it is possible to describe them in the same language. The goal of this study is not to offer concrete solutions to practitioners, but to introduce a high-level view that can help in conceptually transferring methods between different domains.

\section{Common Conceptual Model}
In this section, we describe the conceptual model that can serve as a common framework for leaving the data-centric approach in ML.
We will use the following terms and notations:
\begin{itemize}
\item the ML model $M$
\item the dataset $D$
\item the worldview $W$
\item the constraints $C$
\end{itemize}

\begin{figure}
\includegraphics[width=0.7\textwidth]{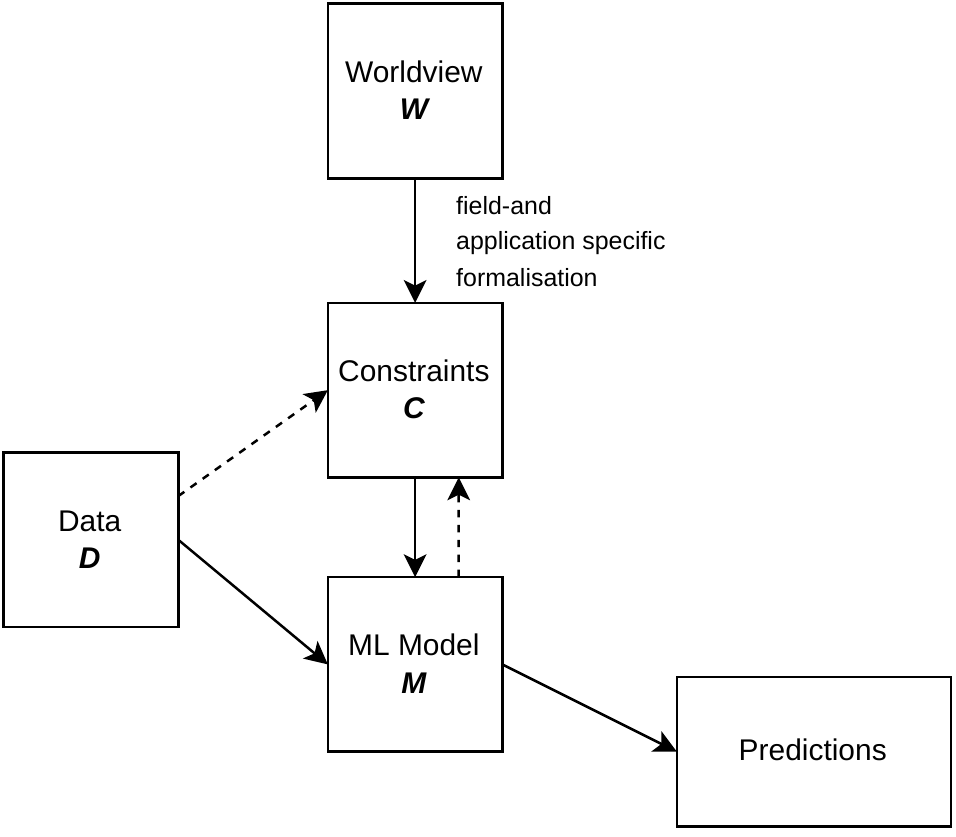}
\caption{\label{fig:conceptual-model}Illustration of the conceptual model: The worldview $W$ forms the basis for the constraints $C$, and is independent of the ML Model $M$ and the data $D$.  $C$ is the translation of $W$ for a given application, data and ML model, and thus also depends on  $D$ and $M$ (indicated by the dotted arrows). The resulting ML Model and predictions are then influenced both by $D$ and $W$.}
\end{figure}

The ML model $M$ is very general here and can be the result of any type  of supervised ML algorithm, including combinations of different algorithms. The ML model $M$ in our discussion also subsumes pre- and post-processing steps. 
The ML model trained with the data $D$ is the raw ML model, which we will denote $M_r$ . If the training algorithm is appropriate, this model will make predictions that adequately represent the relations in $D$. Whether this implies successful generalization, or whether this requires certain train/validation set splits during the learning phase, is immaterial for the remainder of this work, which is why we omit such discussions. 
However, the predictions of $M_r$ need not be \emph{satisfactory} in an actual setting, in which $D$ is only part of the whole thing.
In addition, the worldview $W$ comes in. Our worldview describes how things ought to be, independent of what the data says. The worldview is thus both required and is essential for leaving the data-centric approach. Critically, $W$ can be in stark conflict with $D$. 
$W$ can be broad and generic, such as "A and B have to be treated equally", or "physical law X has to be observed". The formulation of $W$ is not specific to a certain ML model. Instead, for a specific ML model, it must be translated into constraints $C$. The exact form of $C$ will depend on the application and the algorithm used for training. The constraints could either be hard constraints (such as architectural choices, or constraints in classical optimization), or soft constraints (such as regularization, or additional objectives in classical optimization) -- in the latter case one could also refer to them as a "cost". In our discussion, the term constraints $C$ is used in a generic way and covers both types of constraints.
These constraints are then incorporated in the ML Model $M$. This results in the constrained ML Model $M_C$. $M_C$ should then provide \emph{satisfactory} results with respect to $W$. 
The conceptual model is illustrated in fig. \ref{fig:conceptual-model}.
Formalizing $W$, translating it into $C$, and incorporating $C$ into $M$ requires additional knowledge on top of $D$, and the exact process will vary vastly between different fields, applications and algorithms. This is, however, the essential step for leaving the data-centric paradigm.
In the next sections, this will be illustrated with the two fields of physics-informed ML and fair ML.

\section{Examples of applying the conceptual model}
In this section, we first briefly discuss the domains of physics-informed ML and fair ML, and how to approach them with our conceptual model. Then, in less detail, multiple other domains are discussed.
\subsection{$W$ given by physics - physics informed machine learning}
The goal of physics-informed ML (PIML) is to include information about physical laws into ML models. The most widespread form of PIML are  Physics-Informed Neural Networks (PINNs) \cite{raissi_physics-informed_2019}, which incorporate physical laws via sets of ordinary or partial differential equations and which are used to solve initial/boundary value and inverse problems. Here we summarize only the basic concepts of PIML, for a comprehensive overview see \cite{karniadakis_physics-informed_2021}. 
There are two main approaches towards PIML:
\begin{enumerate}
    \item Enforcing physical laws (e.g. certain invariances) with specially designed ML approaches, for example with neural network architectures that automatically follow a certain physical law, or kernels that are given by a certain physical task. Examples include a variant of PINNs, in which initial and boundary values are enforced using predefined functions~\cite{lu2021physics}, or graph neural networks which are trained such that the kernel resembles a physical operator in mesh-based simulations~\cite{pmlr-v119-sanchez-gonzalez20a}.
 \item Including physical laws in the cost function of $M$ (e.g. \cite{sirignano_dgm_2018,erichson_physics-informed_2019,Karpatne_PGNN}). With this, the ML model is penalized in the learning phase if it does not respect the physical law, but it is not actually forced to obey it, and there is no guarantee that it is obeyed. This is a soft approach, in which the physical law(s) are implicitly weighted against other things (e.g. prediction performance). The big upside is that it is generally much easier to include rules/laws in the cost function than enforcing them through specialized architectures/models, and that the same cost function can be used for many different methods.
 \end{enumerate}

In high-level language, in problem settings were PIML is a possible solution we are faced with the following situation: we have data $D$ that approximates the real world, but it is not perfect and can have all types of errors (such as measurement bias, sensor drift, noise, etc). These errors may manifest themselves as  violations of physical laws in $D$, such as an energy loss over time, even if the data is from a system where energy is conserved. If we train an ML model $M_r$ with this raw data only, its predictions will have errors and therefore be unsatisfactory. At least part of these errors will originate from the fact that $D$ has errors. With physics-constrained ML, we incorporate a-priori physical knowledge in order to make the predictions more satisfactory. In this context, $W$ is given by laws of physics, e.g., systems of differential equations or conservation laws that describe that a certain quantity (energy, momentum, mass, etc.) cannot change.
Based on $W$ we thus tell the algorithm “this is how it should be”, because these are the physical laws governing the problem at hand. 
These form the constraints $C$, and making use of the various available methods of PIML, can be used to create the desired constrained model $M_C$.
An illustration of incorporating physics laws in our conceptual model is shown in the left panel of Fig.~\ref{fig:examples}.

\subsection{$W$ given by ethical constraints - fair machine learning}
Another domain where the predictions of the raw model $M_r$ are often unsatisfactory are settings that - either directly or indirectly - involve people. This is the topic of research on algorithmic fairness, and is often alternatively referred to as fair ML, or more colloquially as "bias" problems. In settings with fairness issues, from a  high-level viewpoint, the problem is that “how it should be” is harder to define compared to, e.g., constraints by laws of physics. This is because the worldview $W$, which defines the ethical system that one applies, is not as explicit and quantifiable as a worldview $W$ determined by physical laws. Nevertheless, we can word the problem as “this is how it ought to be, given certain ethical values”, which will form $W$. Then we want the algorithm to follow this worldview $W$. Whether this worldview or the ethical values are prescribed by legal rules or could be seen as a constructed reality does not matter for the algorithm. Just as in physics, where we can say “energy is always conserved”---and if it is not there is a problem in the data and/or the ML model---, in a socio-technical setting we can say “there should not be a difference between different genders with respect to $X$” – and if the ML model does not adhere to this, this means there is a problem in the data and/or the ML model. 
Paralleling the discussion about PIML, also in this setting the data $D$ may not be aligned with $W$. This may be due to selection bias (e.g., some demographic groups are overrepresented in $D$), but may also be caused by relationships in data that are created in conflict with $W$
(for example recruitment datasets that were created by human recruiters, who have biases themselves).  While there are clearly philosophical and practical differences between these issues, and the issues that, e.g., PIML tries to solve, it is still possible to equally describe both problems in the same language with our high-level conceptual model.
\begin{figure}
\includegraphics[width=0.45\textwidth]{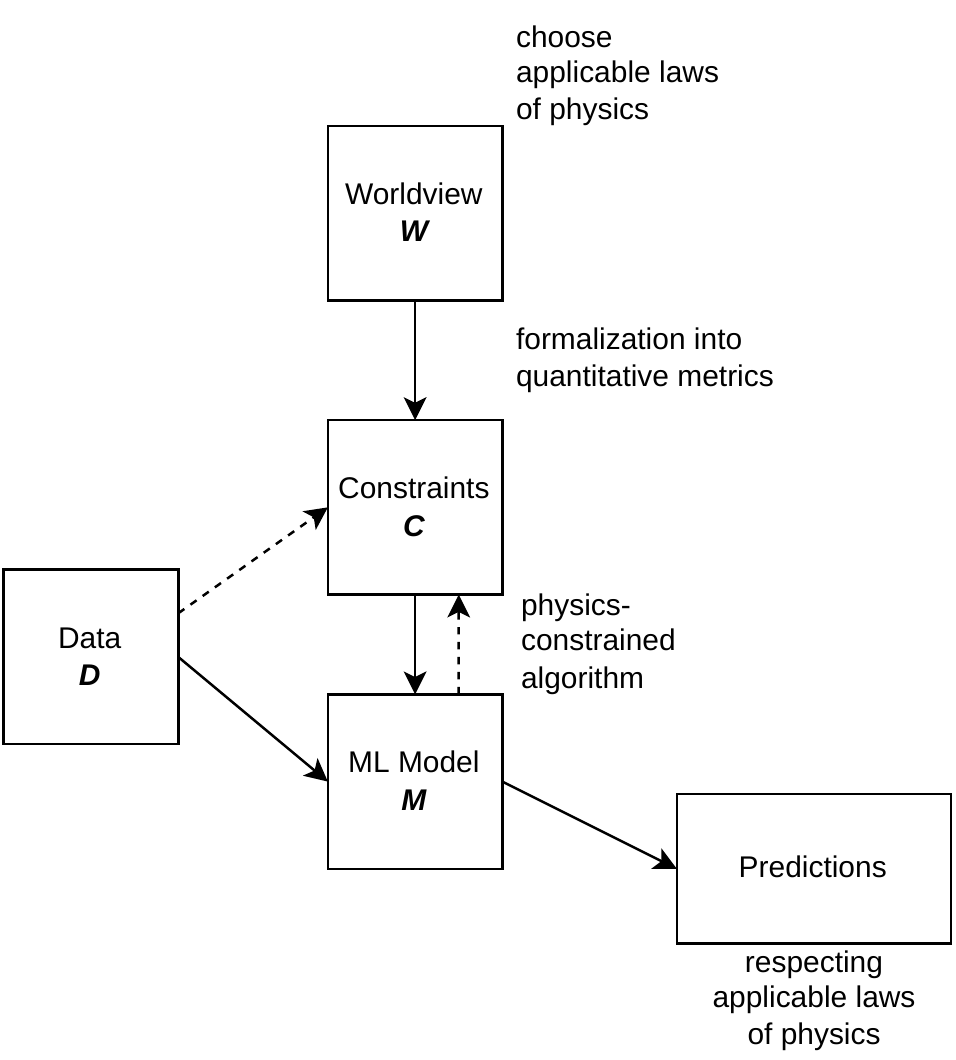}
\includegraphics[width=0.45\textwidth]{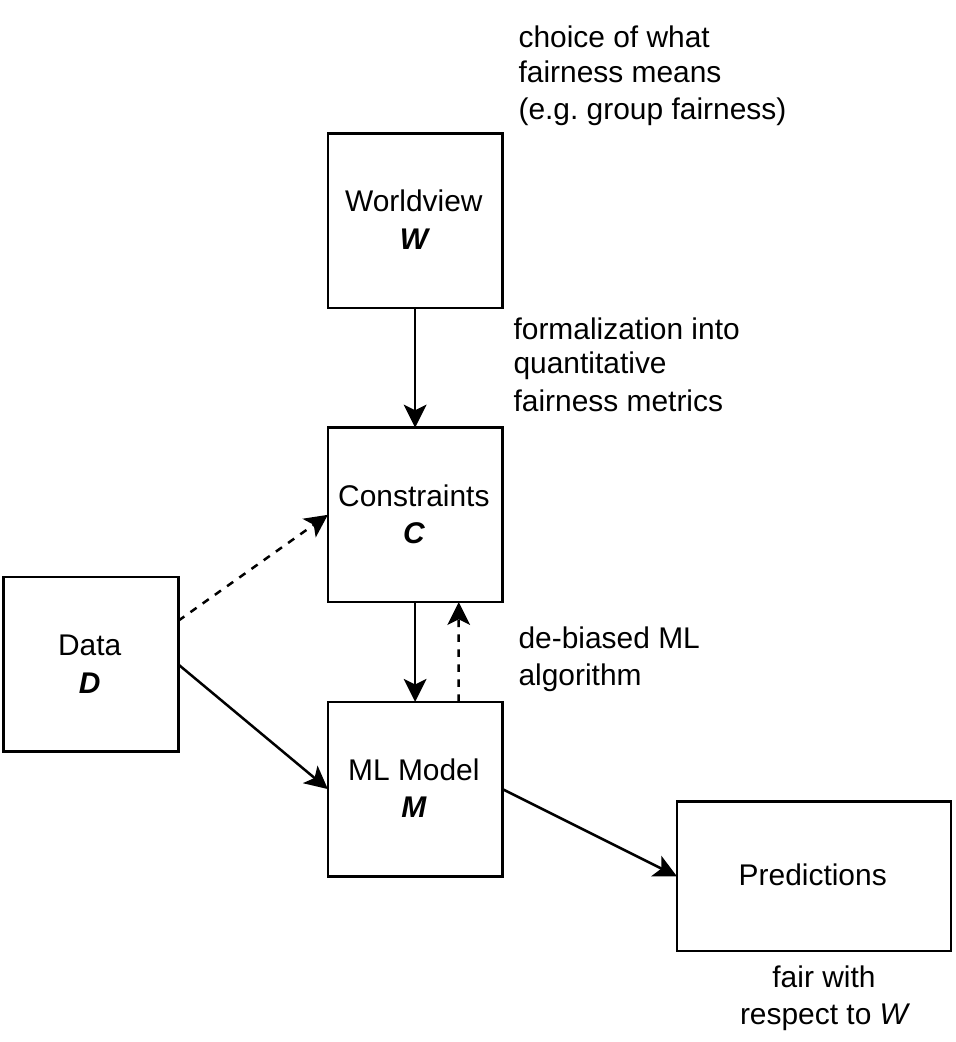}
\caption{\label{fig:examples}Conceptual model applied to physics informed ML (right) and fair ML (left).}
\end{figure}

There is no single definition of fairness in general, and, therefore, also no general definition of a fair ML algorithm (see, e.g., \cite{barocas-hardt-narayanan} for different approaches to fairness metrics). The notion of fairness depends on context (including cultural context) and the  application at hand. However, an important aspect is that different fairness constraints usually contradict each other \cite{barocas-hardt-narayanan}. There are also more complex approach to fairness, for example the question of long-term fairness in changing systems \cite{liu2018delayed,HusseinFairEqual19,damourFairnessNotStatic2020,scher_modelling_2023}.\cite{johnson_towards_2021} gives an overview of work on ethical computing, specifically for software that incorporates data-driven approaches.
\cite{dennerlein_guiding_2020} presents a framework for incorporating ethical principles in technology-enhanced learning software.
IEEE is currently working on establishing a standard for ethically aligned design for autonomous and intelligent systems \cite{aldinhas_ferreira_ieee_2019}
The choice of the fairness definition in our setting is part of our worldview. The exact (quantitative) definition cannot be independent of the used ML model, and is part of translating $W$ to $C$. 

Many different approaches have been proposed for incorporating different fairness constraints into ML models \cite{DBLP:journals/corr/abs-1810-01943,verma2018fairness}. This is often called reducing bias, or  ``debiasing''. Debiasing algorithms either work by modifying the training data,  modifying the training algorithm, post-processing the outputs of the ML model, or are a combination of those approaches.
In our high-level language, all these methods can be incorporated into $M$, while  the chosen fairness definition forms $W$.
The debiasing algorithms then use this definition of fairness as constraint. Some algorithms can work with hard constraints (ensuring that the target fairness is reached), others work with soft constraints (e.g., integrating it in the loss function, in which the training algorithms balances it with other targets such as accuracy).

In contrast to PIML, different fairness definitions usually contradict each other, and we need to \emph{choose} which approach we take. For example, in a setting where fewer individuals from one gender than from another apply to a certain job, it is impossible to meet both the criterion that the same number of individuals of both genders are selected (which would be a valid fairness criterion), and at the same time meet the constraint that the acceptance rates must be the same for both genders (which would also be a valid fairness criterion).

Another important difference between the setting of PIML and the setting of fairness is that the "real world" can be biased, and our beliefs can thus be contradictory to the real world. Take the following example: One might want to prescribe the moral constraint that gender must not play any role in a recruitment process. If a recruitment process is trained on historical data, and recruiters in the past were biased, then the system is trained on data that stems from a process that does not fulfill the fairness constraint that we want to impose. This is not the case for physical constraints. Here of course there can also be errors in the data etc., but fundamentally the underlying system adheres to the constraints. The conceptual model that we present in this paper is still able to describe the two concepts in a single framework. As the choice of fairness approach is part of $W$, we can use the same conceptual model for fairness as well as for PIML. An illustration of incorporating ethical fairness requirements in our conceptual model is shown in the right panel of fig. \ref{fig:examples}.

\subsection{Other domains}
We have exemplarily shown how our conceptual model relates to the fields of fair ML and physics informed ML. However, the problem that models trained on $D$ are not satisfactory for an application, even when the algorithm is appropriate for $D$, does occur also in many other settings. One such area is domain generalization, which describes how well an algorithm that was trained on data from one domain can make predictions on another domain   \cite{blanchard2011generalizing,daomaingen2013,DBLP:journals/corr/abs-2007-01434}. One could argue that this is also a case where without special measures the predictions of the raw ML model are satisfactory on the training domain, but unsatisfactory on the domain it is used in the end. Depending on the exact context, this might or might not be the case: In cases where some data from the target domain is available at training time, the problem of domain generalization would be absorbed into $D$ in our high level view, and it would be down to finding a suitable raw model $M$. However, in cases in which we do not have data from the target domain, but some other information that can - at least in principle - be translated into a constraint, applying our conceptual model makes sense, and might help in adapting solutions from other domains (e.g. fairness) for the specific domain generalization problem one is faced with.

Another related area of research is psychology-informed recommender systems~\cite{lex2021psychology}. This type of recommender system is based on the idea that human decision support should also follow guidelines of human intelligence rather than solely follow guidelines of artificial intelligence (e.g.,~\cite{seitlinger2015attention,ht_ws_2014}). One example is the integration of a human memory model (representing $W$) into recommendation algorithms (representing $M$) to accurately model interest shifts based on the power law of time-dependent decay of item exposure in human memory (representing $C$)~\cite{www_hashtag_2017}.

\section{Conclusion and Outlook}
Data-driven models have enabled breakthroughs and new applications in a wide range of domains. It has, however, also become apparent that in many settings, the models still make predictions that are, in one way or the other, flawed. We referred to such models and their predictions as "unsatisfactory". To improve these unsatisfactory models, many methods have been developed that help to include into the models' knowledge and constraints not present in the training data. In this paper, we have presented a conceptual model that offers a high-level view on how to incorporate additional requirements into data-driven models. We have exemplified the conceptual model on two topics, namely Fair Machine Learning (Fair ML) and Physics-Informed Machine Learning (PIML). When looked at in detail, the two areas both use different methods to tackle different problems. However, with our concept, the approaches can be described in a common language. We hope that this will help in allowing the transfer of ideas between different domains - such as between fair ML and physics informed ML. 
We are convinced that approaching the different domains and areas on such a high-level view will allow identifying the concepts that can be transferred and will help in solving at least some of the open challenges that stem from relying on data-centric approaches.

\begin{acks}
This work was partly supported by the ``DDAI'' COMET Module within
the COMET -- Competence Centers for Excellent Technologies Programme, funded by the Austrian Federal Ministry (BMK and BMDW), the Austrian Research Promotion Agency (FFG), the province of Styria (SFG) and partners from industry and academia. The COMET Programme is managed by FFG.
\end{acks}

\bibliographystyle{ACM-Reference-Format}
\bibliography{manual-references, zotero-references}










\end{document}